\documentclass{bmvc2k}

\usepackage{booktabs}
\usepackage{pifont}

\title{ELDA: Using Edges to Have an Edge on Semantic Segmentation Based UDA}

\addauthor{Ting-Hsuan Liao}{tingforun@gapp.nthu.edu.tw}{*1}
\addauthor{Huang-Ru Liao}{mimiliao2000@gapp.nthu.edu.tw}{*1}
\addauthor{Shan-Ya Yang}{sonya.tw2000@gapp.nthu.edu.tw}{$\dagger$1}
\addauthor{Jie-En Yao}{matt1129yao@gapp.nthu.edu.tw}{$\dagger$1}
\addauthor{Li-Yuan Tsao}{abcdabcd10211@gapp.nthu.edu.tw}{$\dagger$1}
\addauthor{Hsu-Shen Liu}{tenhoshi9023@gapp.nthu.edu.tw}{$\dagger$1}
\addauthor{Bo-Wun Cheng}{bobcheng15@gapp.nthu.edu.tw}{$\ddag$1}
\addauthor{Chen-Hao Chao}{lance_chao@gapp.nthu.edu.tw}{$\ddag$1}
\addauthor{Chia-Che Chang}{chia-che.chang@mediatek.com}{$\ddag$2}
\addauthor{Yi-Chen Lo}{yichen.lo@mediatek.com}{$\ddag$2}
\addauthor{Chun-Yi Lee}{cylee@gapp.nthu.edu.tw}{1}

\addinstitution{
 Elsa Lab, \\ Department of Computer Science, \\ National Tsing Hua University,\\ No.~101, Section~2, Kuang-Fu Road,\\ Hsinchu City 300, \\Taiwan \vspace{0.5em}
}
\addinstitution{
 MediaTek, Inc.,\\
 No.~1, Dusing 1st Road, \\
 Hsinchu Science Park, \\
 Hsinchu City 300, \\
 Taiwan \vspace{0.5em}\\ \hspace{-0.5em}*, $\dagger$, and $\ddag$  indicate equal contributions.
}

\runninghead{Liao ET AL.}{ELDA: Using Edges to Have an Edge on Semantic Segmentation}

\newcommand{\Xs}{X_s}
\newcommand{\Ys}{Y_s}
\newcommand{\Xt}{X_t}

\newcommand{\xs}{x_s}
\newcommand{\ys}{y_s}
\newcommand{\xt}{x_t}
\newcommand{\ypt}{y^{'}_t}

\newcommand{\C}{\text{C}}
\newcommand{\D}{\text{D}}
\newcommand{\CE}{\text{CE}}
\newcommand{\Sig}{\text{Sigmoid}}
\newcommand{\Conv}{\text{Conv}}

\newcommand{\eshatinit}{\hat{e}_s^{\text{init}}}
\newcommand{\ethatinit}{\hat{e}_t^{\text{init}}}
\newcommand{\yshatinit}{\hat{y}_s^{\text{init}}}
\newcommand{\ythatinit}{\hat{y}_t^{\text{init}}}

\newcommand{\eshatfinal}{\hat{e}_s^{\text{final}}}
\newcommand{\ethatfinal}{\hat{e}_t^{\text{final}}}
\newcommand{\yshatfinal}{\hat{y}_s^{\text{final}}}
\newcommand{\ythatfinal}{\hat{y}_t^{\text{final}}}

\newcommand{\fs}{f_{\text{shared}}}
\newcommand{\fy}{f_{\text{seg}}}
\newcommand{\fe}{f_{\text{edge}}}
\newcommand{\fycm}{f_{\text{seg}}^{\text{cm}}}
\newcommand{\fecm}{f_{\text{edge}}^{\text{cm}}}
\newcommand{\fyat}{f_{\text{seg}}^{\text{mid}}}
\newcommand{\feat}{f_{\text{edge}}^{\text{mid}}}

\newcommand{\Lseginit}{L_{\text{seg}}^{\text{init}}}
\newcommand{\Ledgeinit}{L_{\text{edge}}^{\text{init}}}
\newcommand{\Lsegfinal}{L_{\text{seg}}^{\text{final}}}
\newcommand{\Ledgefinal}{L_{\text{edge}}^{\text{final}}}
\newcommand{\Lseg}{L_{\text{seg}}}
\newcommand{\Ledge}{L_{\text{edge}}}
\newcommand{\Ltotal}{L_{\text{total}}}

\begin{document}

\maketitle

\vspace{-3em}
\begin{abstract}
Many unsupervised domain adaptation (UDA) methods have been proposed to bridge the domain gap by utilizing domain invariant information. Most approaches have chosen depth as such information and achieved remarkable successes. Despite their effectiveness, using depth as domain invariant information in UDA tasks may lead to multiple issues, such as excessively high extraction costs and difficulties in achieving a reliable prediction quality. As a result, we introduce Edge Learning based Domain Adaptation (ELDA), a framework which incorporates edge information into its training process to serve as a type of domain invariant information. In our experiments, we quantitatively and qualitatively demonstrate that the incorporation of edge information is indeed beneficial and effective, and enables ELDA to outperform the contemporary state-of-the-art methods on two commonly adopted benchmarks for semantic segmentation based UDA tasks. In addition, we show that ELDA is able to better separate the feature distributions of different classes. We further provide ablation analysis to justify our design decisions.



\end{abstract}

\section{Introduction}
\label{sec::introduction}


Supervised learning for semantic segmentation has achieved unprecedented successes in the past few years. Albeit effective, training a semantic segmentation model in a supervised manner requires pixel-level labeling, which is often prohibitively expensive and time-consuming. Being aware of these undesirable drawbacks, recent studies have been attempting to make use of the existing labeled datasets or simulated environments to train semantic segmentation models, and then adapt the models to certain label-less targeted domains. This category of research direction is often referred to as semantic segmentation based UDA in the computer vision community. A number of approaches have been explored to tackle this challenge, including adversarial training~\cite{vu2019advent,tsai2019domain,xu2019adversarial,wu2020dual}, anchoring~\cite{zhang2019category,zhang2021prototypical,ma2021coarsetofine,ning2021multianchor}, and pseudo labeling (PL)~\cite{zou2018domain,zou2020confidence,mei2020instance,tranheden2020dacs}, and have achieved remarkable adaptation performance. However, they rely solely on the semantic labels in the source domain and the raw input data in the target domain, which limits their performance and thus leaves room for further improvements.


In light of these shortcomings, another branch of work has incorporated domain invariant information into their training processes to help bridge the domain gaps confronted by them. Domain invariant information possess a favorable property: the concept it represents is general across different domains.
This property makes it highly desirable for UDA tasks as it is robust against domain gaps. As a result, such property has inspired researchers to explore the usage of domain invariant information in their UDA methods, in which it is oftentimes embedded into the training objectives of some auxiliary tasks. A commonly adopted type of domain invariant information is depth, which contains clues relating to the distance of the surfaces of scene objects from a viewpoint. A number of methods have been proposed to leverage depth information to help shrink the domain gap~\cite{vu2019dada,Lee2019SPIGANPA,Chen_2019_GIOada,Saha_2021_CVPR_CTRL}. The authors in~\cite{wang2021domain} further utilized self-supervised learning (SSL) to retrieve depth information, with an aim of assisting semantic segmentation based UDA tasks, and achieved remarkable performance. 


Unfortunately, the methods that utilized SSL to retrieve depth information have two crucial constraints: First, the computational cost associated with training an accurate auxiliary SSL-based depth estimation model is often expensive. A few researchers~\cite{wang2021domain} even employed two separate depth estimation models in both the source and the target domains to ensure the quality of the generated depth estimation. This worsens the computational burden incurred, and makes such methods less suitable for real world applications. Second, 
since SSL-based models have no access to ground truth labels, their performance is not comparable to physical sensors (e.g., lidar, stereo camera, etc.) or supervised models~\cite{godard2019digging} in terms of accuracy. In other words, their predictions might deviate from the ground truths, and hence might bring negative impacts on the training process of the semantic segmentation based UDA methods.
 

\begin{figure}[t]
  \centering
  \includegraphics[width=\linewidth]{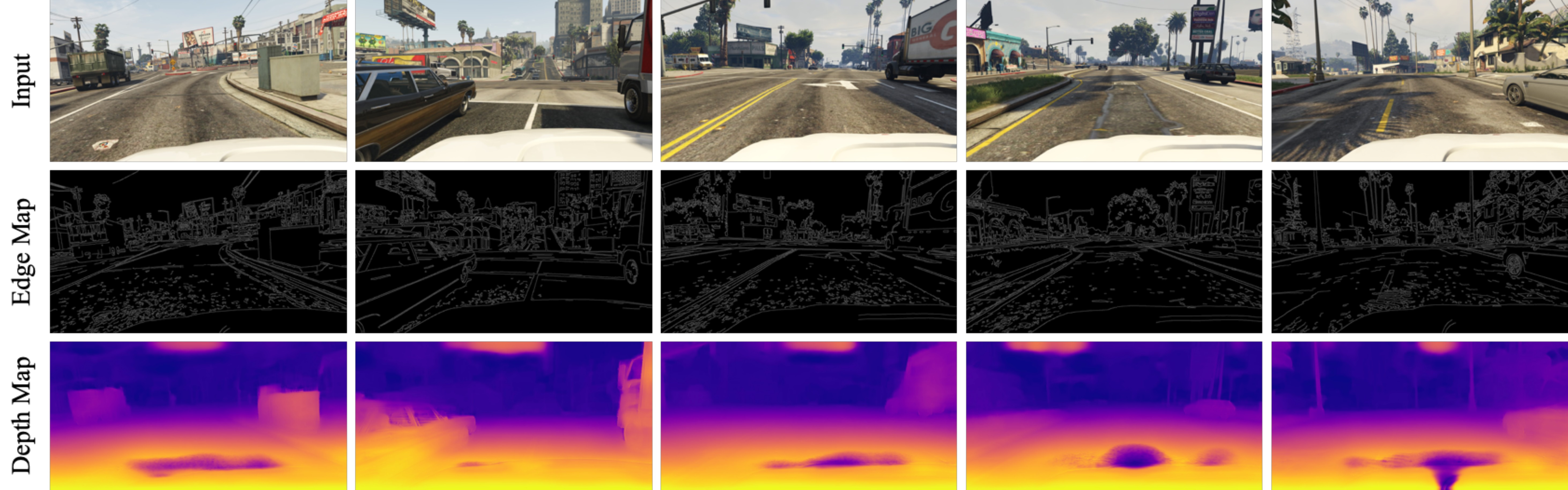}
  \caption{An example that depicts the differences between edges and depth maps extracted by \cite{godard2019digging} on the selected images from the GTA5 dataset \cite{richter2016playing} without the use of any ground truth labels. For the depth maps, nearer surfaces are brighter, while farther ones are darker.}
  \label{fig:depth_vs_edge_example}
\end{figure}

Being aware of the problems associated with using SSL based depth estimation to assist in the training processes of UDA models, we propose to replace it with edges, which is also a type of domain invariant information. The benefits are twofold. First, the computational costs of extracting edges from an input image is substantially lower than those of extracting a depth map from the same image using SSL. Specifically, edges in an image can be obtained by performing convolution using certain fixed kernels over an input image in one pass~\cite{sobel1998}, while the extraction of a depth map usually requires training and inferencing of a sophisticated depth model~\cite{godard2019digging}. Second, the quality of edges is typically much more consistent than that of depth maps, as depth estimation using only RGB images is an ill-posed problem~\cite{eigen2014depth}, and is susceptible to the influences of noises, model architectures, and data distributions. On the contrary, edges are relatively consistent, and less likely to deviate from the ground truth.
Fig.~\ref{fig:depth_vs_edge_example} depicts a motivational example of such characteristics, in which the object boundaries are better captured in the extracted edge maps. In contrast, the depth maps are noisy and the object boundaries are relatively blurred.
The availability of such a high quality boundary information thus offers a promising way to enable a model to better adapt to a target domain.

In order to validate the aforementioned motivation, and take full advantage of the high quality edge information, we propose \textbf{E}dge \textbf{L}earning based \textbf{D}omain \textbf{A}daptation, abbreviated as \textbf{ELDA}. ELDA utilizes edges as the domain invariant information by incorporating edge extraction into its training process as an auxiliary task. The experimental results show that without resorting to ensemble distillation methods \cite{zhang2021prototypical, Chao_2021_CVPR, DBLP:journals/corr/abs-2112-00295} or transformer based architectures \cite{hoyer2021daformer}, ELDA is able to achieve the state-of-the-art performance on two commonly adopted benchmarks~\cite{cordts2016cityscapes,richter2016playing,Ros_2016_CVPR}. 
The contributions of this work are summarized as follows:


\begin{itemize}
\item We introduce the use of edge information as an auxiliary task for semantic segmentation based UDA, and 
develop an effective framework named ELDA, to take full advantage of the valuable edge information embedded in the images of both domains.
\item We validate ELDA on two commonly adopted benchmarks quantitatively and qualitatively, and show that it is able to achieve superior performance to the previous methods.
\item We demonstrate that by incorporating edge information into semantic segmentation based UDA, ELDA can capture fine-grained features in an unlabeled target domain.
\end{itemize}



\section{Related Work}
\label{sec::related_work}
\subsection{Unsupervised Domain Adaptation for Semantic Segmentation} 
\label{subsec:uda_for_seg}

A number of UDA methods have been proposed in an attempt to bridge the discrepancies between different domains. One category of these works adopted adversarial training process to learn representations of their target domains~\cite{vu2019advent,tsai2019domain,xu2019adversarial,wu2020dual,vu2019dada}. These frameworks often consist of a generator and a discriminator trained against each other in order to minimize the domain gap, and have shown significant improvements over those trained in the source domains only without the use of any adaptation technique. Another line of work has focused on
self-training and data augmentation measures to tackle UDA problems. For those utilizing self-training, the concentration was mainly on preventing overfitting by using regularization~\cite{zou2020confidence, zheng2020rectifying} or class-balancing~\cite{mei2020instance, zou2018domain} when minimizing uncertainty in their target domains. The authors in~\cite{tranheden2020dacs} extended the concept of self-training and proposed a data augmentation technique. 
It fine-tunes a model with mixed labels generated by combining ground truth annotations from a source domain and pseudo labels from a target domain. Recent researchers further employed
ensemble learning to deal with the above challenge~\cite{Chao_2021_CVPR}.

\subsection{Auxiliary Tasks in Semantic Segmentation based UDA Methods}
\label{subsec::aux_task_segmentation_UDA}

As mentioned in Section~\ref{sec::introduction}, a few recent researchers have turned their attention to incorporating domain invariant features into the training processes of their UDA methods. To achieve this objective, a commonly adopted method is to introduce them through 
auxiliary tasks. Among these tasks, depth estimation is the most widely used one. The main incentive behind this is that geometric and semantic information are highly correlated~\cite{Chen_2019_GIOada, Lee2019SPIGANPA, vu2019dada, Saha_2021_CVPR_CTRL, wang2021domain, guizilini2021geometric}. This concept is first introduced in SPIGAN~\cite{Lee2019SPIGANPA}, which uses synthetic semantic segmentation and depth information as additional means of regularization for their style transfer model. The authors in DADA~\cite{vu2019dada} proposed to shrink the domain gap by fusing segmentation and depth maps together during the adversarial training process. Moreover, CorDA~\cite{wang2021domain} explicitly exploits a segmentation path and a depth estimation path in their UDA framework, where the latter path serves as an auxiliary task. These two paths are interleaved, in which the embeddings from them are fused together using attention layers such that two paths can mutually benefit from each other. In addition, CorDA leverages the prediction discrepancies from two depth decoders to assist in pseudo-label refinement for the segmentation path. Furthermore, GUDA~\cite{guizilini2021geometric} leverages more auxiliary paths, including depth estimation, surface normal, as well as self-supervised photometric loss, in their UDA model. However, the aforementioned methods require additional depth estimation models trained based on certain self-supervised learning approaches and may suffer from the issues mentioned in Section~\ref{sec::introduction}.


\subsection{Usage of Edge Information in Other Computer Vision Domains}
\label{subsec::edge_infor_in_aux_task}

Edge detection has been utilized in a wide variety of computer vision research domains such as semantic segmentation \cite{chen2016semantic, JSENet}, object detection\cite{ferrari2007groups, lim2013sketch}, facial recognition\cite{Facial}, and representation learning \cite{chen2020simple}. Edges are low-level representations extracted from images that are able to reflect important information about the discontinuities in depth, surface orientation, material properties, as well as scene illumination in images. Besides its abundance in semantic information, edges from an image can be easily extracted using conventional computer vision approaches such as Sobel~\cite{sobel1998} and Canny edge extraction algorithms~\cite{canny1986}. These algorithms are efficient and straightforward and do not require the use of any learnable parameter. Therefore they are considered readily available from almost all image domains, and meet the characteristics of domain invariant features described in Section~\ref{sec::introduction}.  Despite these advantages, the use of edge information has not been explored in semantic segmentation based UDA tasks. In this work, we show that edge information can be used as a type of domain invariant feature and can indeed help boost the performance of the proposed 
ELDA model.

\section{Methodology}
\label{sec::methodology}
\subsection{Problem Formulation}
\label{subsec::problem_formulation}

In UDA tasks, a model has access to a source dataset $\Xs=\{\xs^1, ..., \xs^N\}$, their corresponding labels $\Ys=\{\ys^1, ... , \ys^N\}$, and a target dataset $\Xt = \{\xt^1, ... , \xt^M\}$, where $N$ and $M$ denote the number of instances from the source and target domains, respectively. Specifically, a tuple $(\xs,\ys)$ represents an image-label pair from the source domain, and $\xt$ represents a target domain image. The training objective is to train a model such that its predictions can best estimate the ground truth labels in the target domain. In other words, the mean intersection-over-union (mIoU) of the predictions from the model should be maximized. For the detailed notation used in our work, please refer to the notation table in the supplementary material.

\subsection{Overview of the Proposed ELDA Framework}
\label{subsec::ELDA_framework_overview}

\begin{figure}[t]
  \centering
  \includegraphics[width=\linewidth]{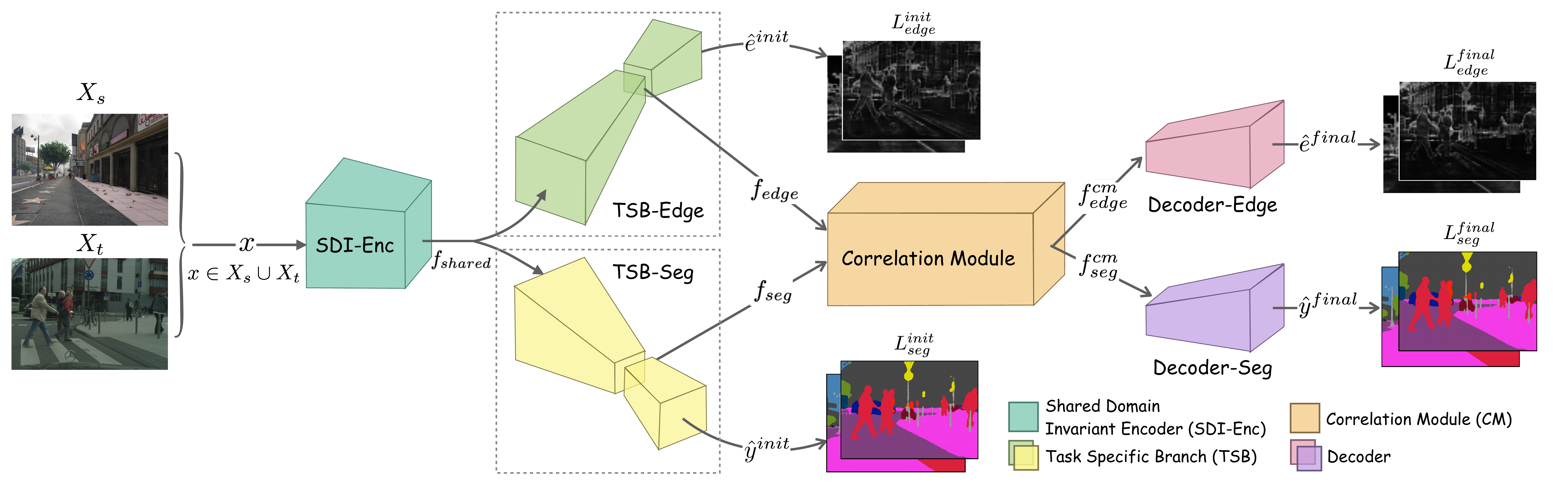}
  \caption{An illustration of the proposed ELDA framework. 
  }
  \label{fig:ELDA_framework_overview}
\end{figure}

Fig.~\ref{fig:ELDA_framework_overview} illustrates an overview of the proposed ELDA framework. First, an input image from either the source domain (i.e., $\xs\in\Xs$) or the target domain (i.e., $\xt\in\Xt$) is first fed into a shared domain invariant encoder (SDI-Enc) to obtain the shared latent feature embedding ($\fs$). Subsequently, $\fs$ is passed through two separate task specific branches (TSBs), to generate the task specific features, i.e., the edge ($\fe$) and the semantic segmentation ($\fy$) features, and the initial predictions of edges ($\eshatinit$ or $\ethatinit$) as well as semantic segmentation ($\yshatinit$ or $\ythatinit$).
Both $\fe$ and $\fy$ are then propagated through a correlation module (CM) to exchange information between $\fe$ and $\fy$. Then, the outputs of CM are forwarded to two distinct decoders to generate the final output predictions of edges ($\eshatfinal$ or $\ethatfinal$) and semantic segmentation ($\yshatfinal$ or $\ythatfinal$), respectively, where subscripts $s$ and $t$ denote the source and the target domains. The edge detection loss $\Ledge$ and the segmentation loss $\Lseg$ are computed to update the model's weights. In the following subsections, we explain the components of ELDA. In Section~\ref{subsec::ELDA_architecture} we elaborate on the details of SDI-Enc, TSBs, and CM. 
In Section~\ref{subsec::loss_function}, we describe the formulations of the loss functions $\Ledge$ and $\Lseg$.





\subsection{Architecture Components of the ELDA Framework}
\label{subsec::ELDA_architecture}


\subsubsection{Shared Domain Invariant Encoder (SDI-Enc)}
\label{subsubsec::SDI-Enc}

In auxiliary task learning, shared encoders are
usually adopted to extract common features so as to enhance performance and reduce inference cost \cite{mtinet}. ELDA employs the shared encoder structure in~\cite{wang2021domain} to extract $\fs$
for capturing both edge and segmentation features.



\subsubsection{Task Specific Branch (TSB)}
\label{subsubsec::TSB}

To enable $\fs$ to be further interpreted into specific feature embeddings that bear edge and semantic segmentation meanings, two separate branches of TSBs, similar to those used in~\cite{mtinet, wang2021domain}, are utilized to generate initial edge and segmentation predictions. The two TSBs contain their separate encoders and decoders. The encoders are in charge of encrypting $\fs$ into task specific features $\fe$ and $\fy$, which are later fed to the CM. Meanwhile, the decoders are employed to decode $\fe$ and $\fy$ in to $\eshatinit$ or $\ethatinit$ and $\yshatinit$ or $\ythatinit$, respectively, depending on the original domain of the input image, for updating SDI-Enc and the TSBs. Please note that $\hat{e}$ represents the edge predictions, $\hat{y}$ denotes the semantic segmentation predictions, and the subscripts $s$ and $t$ stand for the source and the target domains. 

\subsubsection{Correlation Module (CM)}
\label{subsubsec::CM}
With an aim to communicate information between 
$\fe$ and $\fy$, we employ a CM~\cite{padnet, wang2021domain} into ELDA. Specifically, 
within CM, the information in $\fe$ and $\fy$ are first filtered with sigmoid functions to re-weight the task specific intermediate embeddings $\fyat$ and $\feat$ as:
\begin{equation}
\fyat = \Conv(\fy),\,\,\,\,
\feat  = \Conv(\fe),
\end{equation}
\begin{equation}
\fycm = \fy + \feat * \Sig(\Conv(\fe)),
\end{equation}
\begin{equation}
\fecm  = \fe + \fyat * \Sig(\Conv(\fy)),
\end{equation}
where $\Conv(\cdot)$ and $\Sig(\cdot)$ denote the convolution and sigmoid functions, and $\fycm$ and $\fecm$ are the outputs of CM.  CM assists in preserving the essential features from both TSBs. 








\subsection{Loss Function Design}
\label{subsec::loss_function}


\subsubsection{The Loss Function for Edge Estimation ($\Ledge$)}

In ELDA, the supervision targets for edges in both the source and the target domains are generated using the Canny edge extraction algorithm~\cite{canny1986} denoted as $\C(\cdot\,;\sigma)$, where $\sigma$ is the parameter for controlling the smoothness of an edge map. The edge loss $\Ledge = \Ledgeinit +\Ledgefinal$ is then computed between the edge predictions from ELDA and the edges generated by $\C(\cdot\,;\sigma)$. Both $\Ledgeinit$ and $\Ledgefinal$ are derived by extending the DICE loss~\cite{milletari2016vnet}, as it is able to prevent imbalance between different classes.
The expressions of $\Ledgeinit$ and $\Ledgefinal$ are formulated as:
\begin{equation}
    \Ledgeinit = (1 - \D(\C(\xs\,;\sigma), \eshatinit)) + (1 - \D(\C(\xt\,;\sigma), \ethatinit)),
\end{equation}
\begin{equation}
    \Ledgefinal = (1 - \D(\C(\xs\,;\sigma), \eshatfinal)) + (1 - \D(\C(\xt\,;\sigma), \ethatfinal)),
\end{equation}
\begin{equation}
\D(e, \hat{e})= \frac{2 \sum_{i=1}^P e \hat{e}}{\sum_{i=1}^P e^2 + \sum_{i=1}^P \hat{e}^2 },\,\,\,\, 0\leq \D \leq 1,
\end{equation}
where $P$ represents the number of pixels in an input image, $\D(\cdot)$ denotes the DICE loss operator, $e\in\{0,1\}$ represents the edges generated by $\C(\cdot\,;\sigma)$, and $\hat{e}\in[0,1]$ denotes the edges predicted by ELDA. Please note that $\hat{e}$ can be any of  $\eshatinit$, $\ethatinit$, $\eshatfinal$, or $\ethatfinal$.


\subsubsection{The Loss Function for Semantic Segmentation ($\Lseg$)}
In ELDA, the training targets for semantic segmentation in the source domain are the ground truth labels, while those in the target domain are the pseudo labels generated using ELDA's predictions. The loss for semantic segmentation $\Lseg = \Lseginit + \Lsegfinal$ is then computed between the predictions of ELDA and the training targets by utilizing the cross-entropy (CE) operator $\CE(\cdot)$~\cite{DBLP:journals/corr/abs-1805-07836}. The expressions of the loss components $\Lseginit$ and $\Lsegfinal$ are formulated as follows:
\begin{equation}
    \Lsegfinal = \CE(\ys, \yshatfinal)+ \CE(\ypt, \ythatfinal),\,\,\,\,
    \Lseginit = \CE(\ys, \yshatinit)+ \CE(\ypt, \ythatinit),
\end{equation}
where $\CE(y, \hat{y})= - \sum_{i=1}^P y \log \hat{y}$, $\ypt$ represents the pseudo-labels in the target domain, and $y$ denotes the segmentation labels, which can be either $\ys$ or $\ypt$. Finally, $\hat{y}$ denotes the predicted segmentation maps. Please note that $\hat{y}$ can be any of $\yshatinit$, $\ythatinit$,  $\yshatfinal$, or $\ythatfinal$.




\subsubsection{The Total Loss ($\Ltotal$)}

Based on the above derivations, the total loss can be formulated as $\Ltotal = \Lseg + \lambda \Ledge$,
where $\lambda$ is a balancing factor whose value is provided in the supplementary material.

\section{Experimental Results}
\label{sec::experiments}





\subsection{Experimental Setup}
\label{subsec::experimental_setup}

We evaluate and compare the experimental results of ELDA against the pure semantic segmentation based UDA methods~\cite{mei2020instance,tranheden2020dacs,zhang2021prototypical,zhang2019category,zheng2020rectifying,zou2018domain,yang2020fda,lv2020pit}, as well as the methods that take advantage of additional information in the form of auxiliary tasks~\cite{vu2019dada,wang2021domain,Lee2019SPIGANPA,Chen_2019_GIOada,Saha_2021_CVPR_CTRL,guizilini2021geometric}.
The goal of the comparisons against~\cite{tranheden2020dacs}, which is a pure semantic segmentation based UDA method that bears a similar model architecture as ELDA, is to demonstrate that the additional edge information incorporated into the training process of ELDA is indeed beneficial. In addition, the goal of the comparisons against~\cite{wang2021domain}, which is one of the methods that utilize auxiliary tasks and is considered the state-of-the-art, is to show that edges work as well as, or even better than depth, when it comes to designing auxiliary tasks.
We evaluate ELDA and the baselines on two commonly adopted benchmarks: GTA5~\cite{richter2016playing}$\to$ Cityscapes~\cite{cordts2016cityscapes} and SYNTHIA~\cite{Ros_2016_CVPR}$\to$ Cityscapes. 
The detailed setups are offered in the supplementary material.

\begin{table*}[t]
\centering
\resizebox{\textwidth}{!}{%
\renewcommand{\arraystretch}{1.2}
\newcommand{\mytoprule}{\toprule[1.5pt]}
\footnotesize
\begin{tabular}{lllllllllllllllllllll}
\mytoprule
\multicolumn{20}{c}{GTA5 $\to$ Cityscapes} &  \\ \mytoprule
\multicolumn{1}{c|}{Method} & \multicolumn{1}{c}{Road} & \multicolumn{1}{c}{SideW} & \multicolumn{1}{c}{Build} & \multicolumn{1}{c}{Wall} & \multicolumn{1}{c}{Fence} & \multicolumn{1}{c}{Pole} & \multicolumn{1}{c}{Light} & \multicolumn{1}{c}{Sign} & \multicolumn{1}{c}{Veg} & \multicolumn{1}{c}{Terrain} & \multicolumn{1}{c}{Sky} & \multicolumn{1}{c}{Person} & \multicolumn{1}{c}{Rider} & \multicolumn{1}{c}{Car} & \multicolumn{1}{c}{Truck} & \multicolumn{1}{c}{Bus} & \multicolumn{1}{c}{Train} & \multicolumn{1}{c}{Motor} & \multicolumn{1}{c|}{Bike} & \multicolumn{1}{c}{mIoU} \\

\hline
\multicolumn{1}{c|}{Source only} & \multicolumn{1}{c}{70.1} & \multicolumn{1}{c}{18.4} & \multicolumn{1}{c}{66.1} & \multicolumn{1}{c}{12.8} & \multicolumn{1}{c}{17.4} & \multicolumn{1}{c}{22.1} & \multicolumn{1}{c}{30.8} & \multicolumn{1}{c}{16.1} & \multicolumn{1}{c}{79.1} & \multicolumn{1}{c}{14.4} & \multicolumn{1}{c}{71.3} & \multicolumn{1}{c}{57.1} & \multicolumn{1}{c}{23.7} & \multicolumn{1}{c}{77.5} & \multicolumn{1}{c}{29.5} & \multicolumn{1}{c}{37.0} & \multicolumn{1}{c}{4.9} & \multicolumn{1}{c}{29.6} & \multicolumn{1}{c|}{31.5} & \multicolumn{1}{c}{37.3} \\
\hline
\multicolumn{20}{c}{Pure semantic segmentation based UDA methods} &  \\
\hline
\multicolumn{1}{c|}{CBST\cite{zou2018domain}} & \multicolumn{1}{c}{91.8} & \multicolumn{1}{c}{53.5} & \multicolumn{1}{c}{80.5} & \multicolumn{1}{c}{32.7} & \multicolumn{1}{c}{21.0} & \multicolumn{1}{c}{34.0} & \multicolumn{1}{c}{28.9} & \multicolumn{1}{c}{20.4} & \multicolumn{1}{c}{83.9} & \multicolumn{1}{c}{34.2} & \multicolumn{1}{c}{80.9} & \multicolumn{1}{c}{53.1} & \multicolumn{1}{c}{24.0} & \multicolumn{1}{c}{82.7} & \multicolumn{1}{c}{30.3} & \multicolumn{1}{c}{35.9} & \multicolumn{1}{c}{16.0} & \multicolumn{1}{c}{25.9} & \multicolumn{1}{c|}{42.8} & \multicolumn{1}{c}{45.9} \\
\multicolumn{1}{c|}{CAG-UDA\cite{zhang2019category}} & \multicolumn{1}{c}{90.4} & \multicolumn{1}{c}{51.6} & \multicolumn{1}{c}{83.8} & \multicolumn{1}{c}{34.2} & \multicolumn{1}{c}{27.8} & \multicolumn{1}{c}{38.4} & \multicolumn{1}{c}{25.3} & \multicolumn{1}{c}{48.4} & \multicolumn{1}{c}{85.4} & \multicolumn{1}{c}{38.2} & \multicolumn{1}{c}{78.1} & \multicolumn{1}{c}{58.6} & \multicolumn{1}{c}{34.6} & \multicolumn{1}{c}{84.7} & \multicolumn{1}{c}{21.9} & \multicolumn{1}{c}{42.7} & \multicolumn{1}{c}{\textbf{41.1}} & \multicolumn{1}{c}{29.3} & \multicolumn{1}{c|}{37.2} & \multicolumn{1}{c}{50.2} \\
\multicolumn{1}{c|}{FDA\cite{yang2020fda}} & \multicolumn{1}{c}{92.5} & \multicolumn{1}{c}{53.3} & \multicolumn{1}{c}{82.4} & \multicolumn{1}{c}{26.5} & \multicolumn{1}{c}{27.6} & \multicolumn{1}{c}{36.4} & \multicolumn{1}{c}{40.6} & \multicolumn{1}{c}{38.9} & \multicolumn{1}{c}{82.3} & \multicolumn{1}{c}{39.8} & \multicolumn{1}{c}{78.0} & \multicolumn{1}{c}{62.6} & \multicolumn{1}{c}{34.4} & \multicolumn{1}{c}{84.9} & \multicolumn{1}{c}{34.1} & \multicolumn{1}{c}{53.1} & \multicolumn{1}{c}{16.9} & \multicolumn{1}{c}{27.7} & \multicolumn{1}{c|}{46.4} & \multicolumn{1}{c}{50.45} \\
\multicolumn{1}{c|}{PIT\cite{lv2020pit}} & \multicolumn{1}{c}{87.5} & \multicolumn{1}{c}{43.4} & \multicolumn{1}{c}{78.8} & \multicolumn{1}{c}{31.2} & \multicolumn{1}{c}{30.2} & \multicolumn{1}{c}{36.3} & \multicolumn{1}{c}{39.9} & \multicolumn{1}{c}{42.0} & \multicolumn{1}{c}{79.2} & \multicolumn{1}{c}{37.1} & \multicolumn{1}{c}{79.3} & \multicolumn{1}{c}{65.4} & \multicolumn{1}{c}{\textbf{37.5}} & \multicolumn{1}{c}{83.2} & \multicolumn{1}{c}{46.0} & \multicolumn{1}{c}{45.6} & \multicolumn{1}{c}{25.7} & \multicolumn{1}{c}{23.5} & \multicolumn{1}{c|}{49.9} & \multicolumn{1}{c}{50.6}  \\
\multicolumn{1}{c|}{Uncertainty\cite{zheng2020rectifying}} & \multicolumn{1}{c}{90.4} & \multicolumn{1}{c}{31.2} & \multicolumn{1}{c}{85.1} & \multicolumn{1}{c}{36.9} & \multicolumn{1}{c}{25.6} & \multicolumn{1}{c}{37.5} & \multicolumn{1}{c}{\textbf{48.8}} & \multicolumn{1}{c}{48.5} & \multicolumn{1}{c}{85.3} & \multicolumn{1}{c}{34.8} & \multicolumn{1}{c}{81.1} & \multicolumn{1}{c}{64.4} & \multicolumn{1}{c}{36.8} & \multicolumn{1}{c}{86.3} & \multicolumn{1}{c}{34.9} & \multicolumn{1}{c}{52.2} & \multicolumn{1}{c}{1.7} & \multicolumn{1}{c}{29.0} & \multicolumn{1}{c|}{44.6} & \multicolumn{1}{c}{50.3} \\
\multicolumn{1}{c|}{IAST\cite{mei2020instance}} & \multicolumn{1}{c}{93.8} & \multicolumn{1}{c}{57.8} & \multicolumn{1}{c}{85.1} & \multicolumn{1}{c}{39.5} & \multicolumn{1}{c}{26.7} & \multicolumn{1}{c}{26.2} & \multicolumn{1}{c}{43.1} & \multicolumn{1}{c}{34.7} & \multicolumn{1}{c}{84.9} & \multicolumn{1}{c}{32.9} & \multicolumn{1}{c}{88.0} & \multicolumn{1}{c}{62.6} & \multicolumn{1}{c}{29.0} & \multicolumn{1}{c}{87.3} & \multicolumn{1}{c}{39.2} & \multicolumn{1}{c}{49.6} & \multicolumn{1}{c}{23.2} & \multicolumn{1}{c}{34.7} & \multicolumn{1}{c|}{39.6} & \multicolumn{1}{c}{51.5} \\
\multicolumn{1}{c|}{DACS\cite{tranheden2020dacs}} & \multicolumn{1}{c}{89.9} & \multicolumn{1}{c}{39.7} & \multicolumn{1}{c}{87.9} & \multicolumn{1}{c}{30.7} & \multicolumn{1}{c}{39.5} & \multicolumn{1}{c}{38.5} & \multicolumn{1}{c}{46.4} & \multicolumn{1}{c}{52.8} & \multicolumn{1}{c}{88.0} & \multicolumn{1}{c}{44.0} & \multicolumn{1}{c}{88.8} & \multicolumn{1}{c}{\textbf{67.2}} & \multicolumn{1}{c}{35.8} & \multicolumn{1}{c}{84.5} & \multicolumn{1}{c}{45.7} & \multicolumn{1}{c}{50.2} & \multicolumn{1}{c}{0.0} & \multicolumn{1}{c}{27.3} & \multicolumn{1}{c|}{34.0} & \multicolumn{1}{c}{52.1} \\
\multicolumn{1}{c|}{ProDA*\cite{zhang2021prototypical}} & \multicolumn{1}{c}{91.5} & \multicolumn{1}{c}{52.4} & \multicolumn{1}{c}{82.9} & \multicolumn{1}{c}{\textbf{42.0}} & \multicolumn{1}{c}{35.7} & \multicolumn{1}{c}{40.0} & \multicolumn{1}{c}{44.4} & \multicolumn{1}{c}{43.3} & \multicolumn{1}{c}{87.0} & \multicolumn{1}{c}{43.8} & \multicolumn{1}{c}{79.5} & \multicolumn{1}{c}{66.5} & \multicolumn{1}{c}{31.4} & \multicolumn{1}{c}{86.7} & \multicolumn{1}{c}{41.1} & \multicolumn{1}{c}{52.5} & \multicolumn{1}{c}{0.0} & \multicolumn{1}{c}{\textbf{45.4}} & \multicolumn{1}{c|}{53.8} & \multicolumn{1}{c}{53.7} \\

\hline
\multicolumn{20}{c}{Semantic segmentation based UDA methods using auxiliary tasks} &  \\
\hline
\multicolumn{1}{c|}{CorDA\cite{wang2021domain}} & \multicolumn{1}{c}{94.7} & \multicolumn{1}{c}{63.1} & \multicolumn{1}{c}{87.6} & \multicolumn{1}{c}{30.7} & \multicolumn{1}{c}{40.6} & \multicolumn{1}{c}{40.2} & \multicolumn{1}{c}{47.8} & \multicolumn{1}{c}{51.6} & \multicolumn{1}{c}{87.6} & \multicolumn{1}{c}{47.0} & \multicolumn{1}{c}{\textbf{89.7}} & \multicolumn{1}{c}{66.7} & \multicolumn{1}{c}{35.9} & \multicolumn{1}{c}{\textbf{90.2}} & \multicolumn{1}{c}{48.9} & \multicolumn{1}{c}{\textbf{57.5}} & \multicolumn{1}{c}{0.0} & \multicolumn{1}{c}{39.8} & \multicolumn{1}{l|}{\textbf{56.0}} & \multicolumn{1}{c}{56.6} \\
\multicolumn{1}{c|}{ELDA (Ours)} & \multicolumn{1}{c}{\textbf{94.9}} & \multicolumn{1}{c}{\textbf{64.1}} & \multicolumn{1}{c}{\textbf{88.2}} & \multicolumn{1}{c}{35.0} & \multicolumn{1}{c}{\textbf{44.7}} & \multicolumn{1}{c}{\textbf{40.3}} & \multicolumn{1}{c}{47.0} & \multicolumn{1}{c}{\textbf{54.6}} & \multicolumn{1}{c}{\textbf{88.7}} & \multicolumn{1}{c}{\textbf{47.4}} & \multicolumn{1}{c}{88.9} & \multicolumn{1}{c}{67.0} & \multicolumn{1}{c}{31.1} & \multicolumn{1}{c}{{\textbf{90.2}}} & \multicolumn{1}{c}{\textbf{53.7}} & \multicolumn{1}{c}{56.0} & \multicolumn{1}{c}{0.0} & \multicolumn{1}{c}{41.7} & \multicolumn{1}{l|}{55.5} & \multicolumn{1}{c}{\textbf{57.3}} \\
\mytoprule
\end{tabular}}
\caption{The quantitative results evaluated on the GTA5$\to$Cityscapes UDA benchmark. Please note that the distillation stage of ProDA~\cite{zhang2021prototypical} is removed for a fair comparison.
}
\label{tab:gta_benchmark}
\end{table*}
\begin{table*}[t]
\centering
\resizebox{\textwidth}{!}{%
\renewcommand{\arraystretch}{1.2}
\newcommand{\mytoprule}{\toprule[1.5pt]}
\footnotesize
\begin{tabular}{llllllllllllllllllll}
\mytoprule
\multicolumn{18}{c}{SYNTHIA $\to$ Cityscapes} &  \\ \mytoprule
\multicolumn{1}{c|}{Method} & \multicolumn{1}{c}{Road} & \multicolumn{1}{c}{SideW} & \multicolumn{1}{c}{Build} & \multicolumn{1}{c}{Wall} & \multicolumn{1}{c}{Fence} & \multicolumn{1}{c}{Pole} & \multicolumn{1}{c}{Light} & \multicolumn{1}{c}{Sign} & \multicolumn{1}{c}{Veg} & \multicolumn{1}{c}{Sky} & \multicolumn{1}{c}{Person} & \multicolumn{1}{c}{Rider} & \multicolumn{1}{c}{Car} & \multicolumn{1}{c}{Bus} & \multicolumn{1}{c}{Motor} & \multicolumn{1}{c|}{Bike} & \multicolumn{1}{c}{mIoU} \\ 
\hline
\multicolumn{1}{c|}{Source only} & \multicolumn{1}{c}{51.8} & \multicolumn{1}{c}{17.0} & \multicolumn{1}{c}{73.0} & \multicolumn{1}{c}{7.1} & \multicolumn{1}{c}{0.2} & \multicolumn{1}{c}{25.4} & \multicolumn{1}{c}{9.4} & \multicolumn{1}{c}{10.2} & \multicolumn{1}{c}{70.7} & \multicolumn{1}{c}{84.0} & \multicolumn{1}{c}{55.6} & \multicolumn{1}{c}{13.7} & \multicolumn{1}{c}{68.0} & \multicolumn{1}{c}{2.9} & \multicolumn{1}{c}{8.5} & \multicolumn{1}{c|}{16.1} &  \multicolumn{1}{c}{32.1} \\
\hline
\multicolumn{18}{c}{Pure semantic segmentation based UDA methods} &  \\
\hline
\multicolumn{1}{c|}{CBST\cite{zou2018domain}} & \multicolumn{1}{c}{68.0} & \multicolumn{1}{c}{29.9} & \multicolumn{1}{c}{76.3} & \multicolumn{1}{c}{10.8} & \multicolumn{1}{c}{1.4} & \multicolumn{1}{c}{33.9} & \multicolumn{1}{c}{22.8} & \multicolumn{1}{c}{29.5} & \multicolumn{1}{c}{77.6} & \multicolumn{1}{c}{78.3} & \multicolumn{1}{c}{60.6} & \multicolumn{1}{c}{28.3} & \multicolumn{1}{c}{81.6} & \multicolumn{1}{c}{23.5} & \multicolumn{1}{c}{18.8} & \multicolumn{1}{c|}{39.8} & \multicolumn{1}{c}{42.6} \\
\multicolumn{1}{c|}{CAG-UDA\cite{zhang2019category}} & \multicolumn{1}{c}{84.7} & \multicolumn{1}{c}{40.8} & \multicolumn{1}{c}{81.7} & \multicolumn{1}{c}{7.8} & \multicolumn{1}{c}{0.0} & \multicolumn{1}{c}{35.1} & \multicolumn{1}{c}{13.3} & \multicolumn{1}{c}{22.7} & \multicolumn{1}{c}{84.5} & \multicolumn{1}{c}{77.6} & \multicolumn{1}{c}{64.2} & \multicolumn{1}{c}{27.8} & \multicolumn{1}{c}{80.9} & \multicolumn{1}{c}{19.7} & \multicolumn{1}{c}{22.7} & \multicolumn{1}{c|}{48.3} &  \multicolumn{1}{c}{44.5} \\
\multicolumn{1}{c|}{PIT\cite{lv2020pit}} & \multicolumn{1}{c}{83.1} & \multicolumn{1}{c}{27.6} & \multicolumn{1}{c}{81.5} & \multicolumn{1}{c}{8.9} & \multicolumn{1}{c}{0.3} & \multicolumn{1}{c}{21.8} & \multicolumn{1}{c}{26.4} & \multicolumn{1}{c}{33.8} & \multicolumn{1}{c}{76.4} & \multicolumn{1}{c}{78.8} & \multicolumn{1}{c}{64.2} & \multicolumn{1}{c}{27.6} & \multicolumn{1}{c}{79.6} & \multicolumn{1}{c}{31.2} & \multicolumn{1}{c}{31.0} & \multicolumn{1}{c|}{31.3} &  \multicolumn{1}{c}{44.0} \\
\multicolumn{1}{c|}{Uncertainty\cite{zheng2020rectifying}} & \multicolumn{1}{c}{87.6} & \multicolumn{1}{c}{41.9} & \multicolumn{1}{c}{83.1} & \multicolumn{1}{c}{14.7} & \multicolumn{1}{c}{1.7} & \multicolumn{1}{c}{36.2} & \multicolumn{1}{c}{31.3} & \multicolumn{1}{c}{19.9} & \multicolumn{1}{c}{81.6} & \multicolumn{1}{c}{80.6} & \multicolumn{1}{c}{63.0} & \multicolumn{1}{c}{21.8} & \multicolumn{1}{c}{86.2} & \multicolumn{1}{c}{40.7} & \multicolumn{1}{c}{23.6} & \multicolumn{1}{c|}{53.1} &  \multicolumn{1}{c}{47.9} \\
\multicolumn{1}{c|}{IAST\cite{mei2020instance}} & \multicolumn{1}{c}{81.9} & \multicolumn{1}{c}{41.5} & \multicolumn{1}{c}{83.3} & \multicolumn{1}{c}{17.7} & \multicolumn{1}{c}{4.6} & \multicolumn{1}{c}{32.3} & \multicolumn{1}{c}{30.9} & \multicolumn{1}{c}{28.8} & \multicolumn{1}{c}{83.4} & \multicolumn{1}{c}{85.0} & \multicolumn{1}{c}{65.5} & \multicolumn{1}{c}{30.8} & \multicolumn{1}{c}{86.5} & \multicolumn{1}{c}{38.2} & \multicolumn{1}{c}{\textbf{33.1}} & \multicolumn{1}{c|}{52.7} &  \multicolumn{1}{c}{49.8} \\
\multicolumn{1}{c|}{DACS\cite{tranheden2020dacs}} & \multicolumn{1}{c}{80.6} & \multicolumn{1}{c}{25.1} & \multicolumn{1}{c}{81.9} & \multicolumn{1}{c}{21.5} & \multicolumn{1}{c}{2.9} & \multicolumn{1}{c}{37.2} & \multicolumn{1}{c}{22.7} & \multicolumn{1}{c}{24.0} & \multicolumn{1}{c}{83.7} & \multicolumn{1}{c}{90.8} & \multicolumn{1}{c}{67.6} & \multicolumn{1}{c}{38.3} & \multicolumn{1}{c}{82.9} & \multicolumn{1}{c}{38.9} & \multicolumn{1}{c}{28.5} & \multicolumn{1}{c|}{47.6} &  \multicolumn{1}{c}{48.3} \\

\hline
\multicolumn{18}{c}{Semantic segmentation based UDA methods using auxiliary tasks} &  \\
\hline
\multicolumn{1}{c|}{SPIGAN\cite{Lee2019SPIGANPA}} & \multicolumn{1}{c}{71.1} & \multicolumn{1}{c}{29.8} & \multicolumn{1}{c}{71.4} & \multicolumn{1}{c}{3.7} & \multicolumn{1}{c}{0.3} & \multicolumn{1}{c}{33.2} & \multicolumn{1}{c}{6.4} & \multicolumn{1}{c}{15.6} & \multicolumn{1}{c}{81.2} & \multicolumn{1}{c}{78.9} & \multicolumn{1}{c}{52.7} & \multicolumn{1}{c}{13.1} & \multicolumn{1}{c}{75.9} & \multicolumn{1}{c}{25.5} & \multicolumn{1}{c}{10.0} & \multicolumn{1}{c|}{20.5} &  \multicolumn{1}{c}{36.8} \\
\multicolumn{1}{c|}{GIO-Ada\cite{Chen_2019_GIOada}} & \multicolumn{1}{c}{78.3} & \multicolumn{1}{c}{29.2} & \multicolumn{1}{c}{76.9} & \multicolumn{1}{c}{11.4} & \multicolumn{1}{c}{0.3} & \multicolumn{1}{c}{26.5} & \multicolumn{1}{c}{10.8} & \multicolumn{1}{c}{17.2} & \multicolumn{1}{c}{81.7} & \multicolumn{1}{c}{81.9} & \multicolumn{1}{c}{45.8} & \multicolumn{1}{c}{15.4} & \multicolumn{1}{c}{68.0} & \multicolumn{1}{c}{15.9} & \multicolumn{1}{c}{7.5} & \multicolumn{1}{c|}{30.4} &  \multicolumn{1}{c}{37.3} \\
\multicolumn{1}{c|}{DADA\cite{vu2019dada}} & \multicolumn{1}{c}{89.2} & \multicolumn{1}{c}{44.8} & \multicolumn{1}{c}{81.4} & \multicolumn{1}{c}{6.8} & \multicolumn{1}{c}{0.3} & \multicolumn{1}{c}{26.2} & \multicolumn{1}{c}{8.6} & \multicolumn{1}{c}{11.1} & \multicolumn{1}{c}{81.8} & \multicolumn{1}{c}{84.0} & \multicolumn{1}{c}{54.7} & \multicolumn{1}{c}{19.3} & \multicolumn{1}{c}{79.7} & \multicolumn{1}{c}{40.7} & \multicolumn{1}{c}{14.0} & \multicolumn{1}{c|}{38.8} &  \multicolumn{1}{c}{42.6} \\
\multicolumn{1}{c|}{GUDA\cite{guizilini2021geometric}} & \multicolumn{1}{c}{88.1} & \multicolumn{1}{c}{53.0} & \multicolumn{1}{c}{84.0} & \multicolumn{1}{c}{22.0} & \multicolumn{1}{c}{1.4} & \multicolumn{1}{c}{\textbf{39.6}} & \multicolumn{1}{c}{28.2} & \multicolumn{1}{c}{24.8} & \multicolumn{1}{c}{82.7} & \multicolumn{1}{c}{81.5} & \multicolumn{1}{c}{65.5} & \multicolumn{1}{c}{22.7} & \multicolumn{1}{c}{\textbf{89.3}} & \multicolumn{1}{c}{\textbf{50.5}} & \multicolumn{1}{c}{25.1} & \multicolumn{1}{c|}{\textbf{57.5}} &  \multicolumn{1}{c}{51.0} \\
\multicolumn{1}{c|}{CorDA\cite{wang2021domain}} & \multicolumn{1}{c}{\textbf{93.3}} & \multicolumn{1}{c}{\textbf{61.6}} & \multicolumn{1}{c}{85.3} & \multicolumn{1}{c}{19.6} & \multicolumn{1}{c}{\textbf{5.1}} & \multicolumn{1}{c}{37.8} & \multicolumn{1}{c}{36.6} & \multicolumn{1}{c}{42.8} & \multicolumn{1}{c}{84.9} & \multicolumn{1}{c}{90.4} & \multicolumn{1}{c}{69.7} & \multicolumn{1}{c}{41.8} & \multicolumn{1}{c}{85.6} & \multicolumn{1}{c}{38.4} & \multicolumn{1}{c}{32.6} & \multicolumn{1}{c|}{53.9} &  \multicolumn{1}{c}{55.0} \\
\multicolumn{1}{c|}{CTRL\cite{Saha_2021_CVPR_CTRL}} & \multicolumn{1}{c}{86.4} & \multicolumn{1}{c}{42.5} & \multicolumn{1}{c}{80.4} & \multicolumn{1}{c}{20.0} & \multicolumn{1}{c}{1.0} & \multicolumn{1}{c}{27.7} & \multicolumn{1}{c}{10.5} & \multicolumn{1}{c}{13.3} & \multicolumn{1}{c}{80.6} & \multicolumn{1}{c}{82.6} & \multicolumn{1}{c}{61.0} & \multicolumn{1}{c}{23.7} & \multicolumn{1}{c}{81.8} & \multicolumn{1}{c}{42.9} & \multicolumn{1}{c}{21.0} & \multicolumn{1}{c|}{44.7} &  \multicolumn{1}{c}{45.0} \\
\multicolumn{1}{c|}{ELDA (Ours)} & \multicolumn{1}{c}{92.6} & \multicolumn{1}{c}{56.6} & \multicolumn{1}{c}{\textbf{85.5}} & \multicolumn{1}{c}{\textbf{24.2}} & \multicolumn{1}{c}{2.1} & \multicolumn{1}{c}{37.6} & \multicolumn{1}{c}{\textbf{38.1}} & \multicolumn{1}{c}{\textbf{43.1}} & \multicolumn{1}{c}{\textbf{85.7}} & \multicolumn{1}{c}{\textbf{91.5}} & \multicolumn{1}{c}{\textbf{69.8}} & \multicolumn{1}{c}{\textbf{42.0}} & \multicolumn{1}{c}{87.2} & \multicolumn{1}{c}{47.6} & \multicolumn{1}{c}{20.0} & \multicolumn{1}{c|}{50.1} & \multicolumn{1}{c}{\textbf{55.2}} \\
\mytoprule
\end{tabular}}
\caption{The quantitative results evaluated on the SYNTHIA$\to$Cityscapes UDA benchmark.}
\label{tab:syn_benchmark}
\end{table*}

\subsection{Quantitative Results on the Benchmarks}
\label{subsec::quantitative_results}
Tables~\ref{tab:gta_benchmark} and~\ref{tab:syn_benchmark} compare the results of ELDA against multiple baselines. Please note that these baselines do not include works that resort to ensemble distillation methods~\cite{zhang2021prototypical,Chao_2021_CVPR,DBLP:journals/corr/abs-2112-00295} or transformer based architectures~\cite{hoyer2021daformer} for fair comparisons. We also include the performance of ELDA trained solely in the source domains, denoted as \textit{source only}, for reference.
Table~\ref{tab:gta_benchmark} reports the results evaluated on the GTA5$\to$Cityscapes benchmark. By leveraging edge prediction as an auxiliary task, ELDA achieves an mIoU of 57.3\%, outperforming \textit{source only} by a margin of 20\% mIOU. This demonstrates that the addition of edge information brings positive influences. Furthermore, ELDA indeed has an edge over DACS~\cite{tranheden2020dacs} and the current state-of-the-art CorDA~\cite{wang2021domain} by 5.2\% mIoU and 0.7\% mIoU, respectively. Table~\ref{tab:syn_benchmark} shows the results evaluated on the SYNTHIA$\to$Cityscapes benchmark. The results display a similar trend as those on the GTA5$\to$Cityscapes benchmark, in which ELDA is able to reach the state-of-the-art performance of 55.2\% mIOU, outperforming all the baselines. Please note that in this benchmark, the depth ground truth labels in the source domain are provided, and are leveraged by a few baselines~\cite{vu2019dada,Lee2019SPIGANPA,Chen_2019_GIOada,Saha_2021_CVPR_CTRL,guizilini2021geometric,wang2021domain} in their auxiliary tasks. In contrast, ELDA is able to achieve superior performance without the use of any extra labeled data.

\begin{figure}[t]
  \centering
  \includegraphics[width=\linewidth]{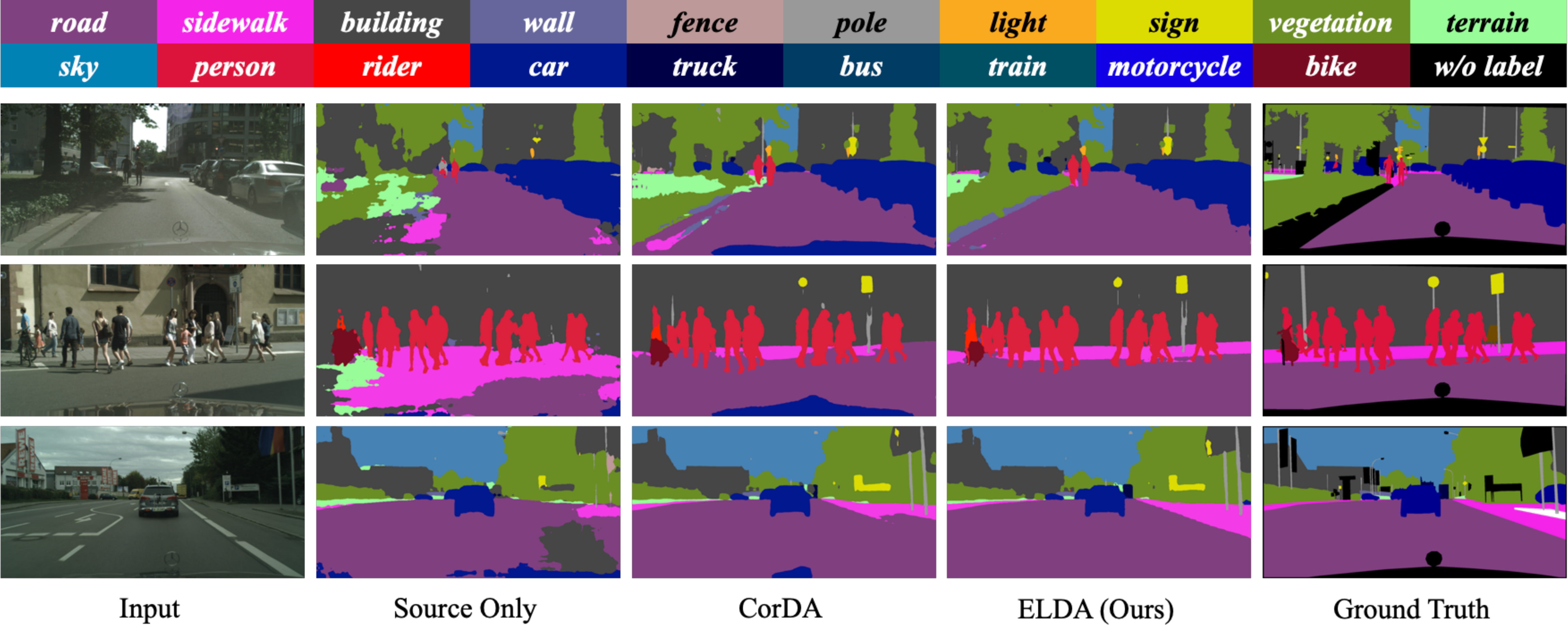}
  \caption{A comparison of the semantic segmentation predictions from different methods. 
  }
  \label{fig:Seg_Qualitative_Result}
\end{figure}

\begin{figure}[t]
  \centering
  \includegraphics[width=\linewidth]{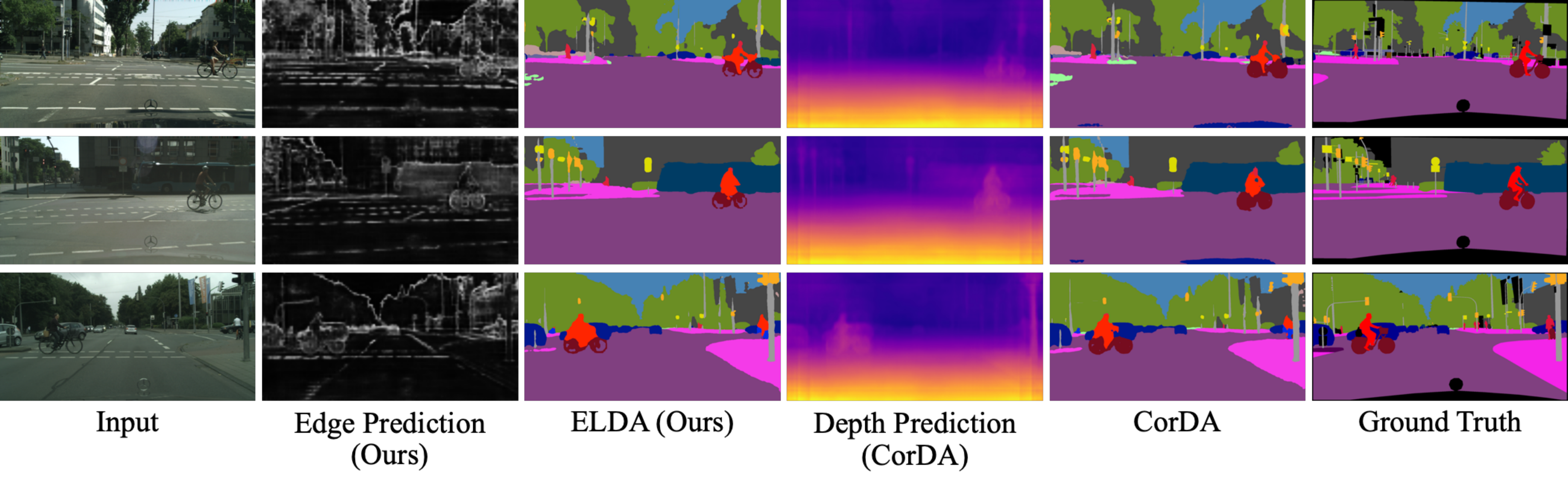}
  \caption{
  A comparison of the predictions evaluated by ELDA and CorDA.
  }
  \label{fig:Edge_Qualitative_Result}
\end{figure}

\subsection{Qualitative Results on the GTA5$\to$Cityscapes benchmark}
\label{subsec::qualitative_result}

Fig.~\ref{fig:Seg_Qualitative_Result} presents the predictions from \textit{source only}, CorDA~\cite{wang2021domain}, and ELDA on some images selected from the GTA5$\to$Cityscapes benchmark. It is observed that predictions from ELDA are of better quality as they are less fragmented and have clearer boundaries as compared to those predicted by \textit{source only} and CorDA. For example, ELDA is able to clearly mark the boundaries between sidewalks and roads, while CorDA and \textit{source only} tend to produce uncertain and sub-optimal boundaries.  Fig.~\ref{fig:Edge_Qualitative_Result} further shows how ELDA is able to produce impressive details in its predictions through approximating the training target of edges generated by $\C(\cdot\,;\sigma)$. The incorporation of high quality edge information allows ELDA to even capture small and subtle features of the image, like the hollowness at the center of bike wheels. In contrast, the predictions from CorDA fail to capture a similar degree of details, which is due to the use of relatively ambiguous depth information during its training process.



\subsection{Ablation Study}
\label{subsec::ablation_study}


\begin{table*}[t]
\centering
\setlength{\tabcolsep}{1em}
\resizebox{\textwidth}{!}{%
\renewcommand{\arraystretch}{1.3}
\newcommand{\mytoprule}{\toprule[1.5pt]}
\footnotesize
\begin{tabular}{l|c|c|c}
\mytoprule
Method & Edge Aux. & Correlation Module & mIoU \\
\hline
DACS \cite{tranheden2020dacs} &&& 52.1 \\
ELDA (SDI-Enc+TSB) & \ding{52} && 55.5 \\
ELDA (SDI-Enc+TSB+CM) & \ding{52} & \ding{52} & \textbf{57.3} \\
\mytoprule
\end{tabular}}
\caption{An ablation analysis for validating the effectiveness of each component.}
\label{tab:component}
\end{table*}

As described in Section~\ref{sec::methodology}, ELDA is made up of multiple key components such as SDI-Enc, TSB, and CM. To examine how each of them contributes to the overall performance, we compare the performances between (1)~ELDA, (2)~ELDA without CM (denoted as ELDA~(SDI-Enc+TSB)), and (3)~ELDA without SDI-Enc, TSB, and CM (which is essentially the DACS baseline~\cite{tranheden2020dacs}). 
The experimental results presented in Table~\ref{tab:component} show that with the addition of each component, the performance grows, indicating that the adoption of these components in ELDA indeed has positive performance impacts and thus validates our design choices. 

\subsection{Visualization of the Features in the Target Domain}
\label{subsec::visualization}

\begin{figure}[t]
  \centering
  \includegraphics[width=\linewidth]{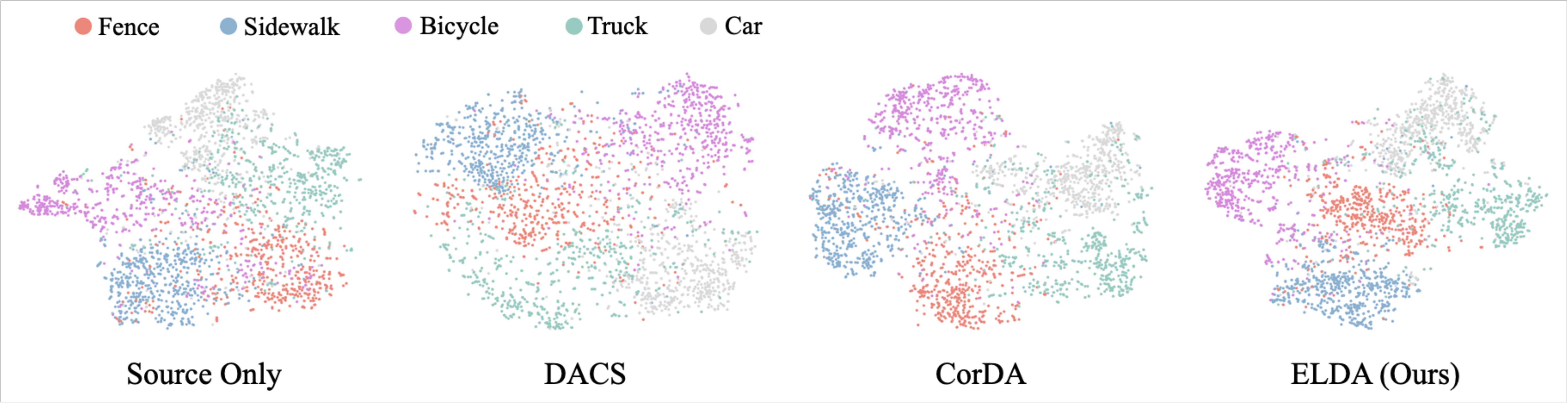}
  \caption{
  A visualization of the feature space using t-SNE for four semantic classes.
  }
  \label{fig:TSNE}
\end{figure}

To verify that ELDA is able to learn a more discriminative and thus descriptive feature space, we visualize the target domain features of \textit{source only}, DACS, CorDA, and ELDA in Fig.~\ref{fig:TSNE} using t-SNE~\cite{vanDerMaaten2008}. Please note that for  ease of viewing and observation, we only display the target domain features of several semantic classes, including sidewalk, fence, car, truck, and bicycle. It is observed that the target domain features captured by~\textit{source only} does not have clear boundaries between different classes, reflecting its sub-optimal performance in the target domain. 
In contrast, the features captured by ELDA, DACS~\cite{tranheden2020dacs}, and CorDA~\cite{wang2021domain} display 
clearer boundaries, translating to their superior performance in the target domain.



\section{Conclusion}
\label{sec::conclusions}

In this work, we proposed an effective framework, called ELDA, for performing semantic segmentation based UDA. ELDA utilizes the highly available and high quality edge information as the domain invariant information by incorporating edge extraction into its training process as an auxiliary task.
To validate the performance of ELDA, we evaluated it against a number of baselines on two commonly adopted benchmarks, and quantitatively and qualitatively showed that ELDA is able to achieve the state-of-the-art performance as compared to the baselines. In addition, we presented an ablation study and feature analysis in the target domain to validate our design choices. As the proposed framework is able to leverage precious edge information to enhance its adaptation performance, it thus offers a different direction to further enhance the performance of semantic segmentation based UDA models.



\section{Acknowledgments}
\label{sec::acknowledgments}
This work was supported by the National Science and Technology Council (NSTC) in Taiwan under grant numbers MOST 111-2223-E-007-004-MY3 and MOST 111-2628-E-007-010. The authors acknowledge the financial support from MediaTek Inc., Hsichu City, Taiwan. The authors would also like to acknowledge the donation of the GPUs from NVIDIA Corporation and NVIDIA AI Technology Center (NVAITC) used in this research work. The authors thank National Center for High-Performance Computing (NCHC) for providing computational and storage resources. 
Finally, the authors would also like to thank the time and effort of the anonymous reviewers for reviewing this paper.

\bibliography{egbib} 
\end{document}